# CONTENT BASED IMAGE RETRIEVAL USING EXACT LEGENDRE MOMENTS AND SUPPORT VECTOR MACHINE

Ch.Srinivasa Rao[1] , S.Srinivas Kumar[2] and B.Chandra Mohan[3]

[1]Department of ECE , Sri Sai Aditya Institute of Science & Technology, Surampalem, A.P, India
*ch_rao@rediffmail.com*
[2]Department of ECE, UCE, JNTUK, Kakinada, A.P, India
*samay_ssk2@yahoo.com*
[3]Department of ECE, Bapatla Engineering College, Bapatla, A.P, India
*chandrabhuma@gmail.com*

## ABSTRACT

*Content Based Image Retrieval (CBIR) systems based on shape using invariant image moments, viz., Moment Invariants (MI) and Zernike Moments (ZM) are available in the literature. MI and ZM are good at representing the shape features of an image. However, non-orthogonality of MI and poor reconstruction of ZM restrict their application in CBIR. Therefore, an efficient and orthogonal moment based CBIR system is needed. Legendre Moments (LM) are orthogonal, computationally faster, and can represent image shape features compactly. CBIR system using Exact Legendre Moments (ELM) for gray scale images is proposed in this work. Superiority of the proposed CBIR system is observed over other moment based methods, viz., MI and ZM in terms of retrieval efficiency and retrieval time. Further, the classification efficiency is improved by employing Support Vector Machine (SVM) classifier. Improved retrieval results are obtained over existing CBIR algorithm based on Stacked Euler Vector (SERVE) combined with Modified Moment Invariants (MMI).*

## 1. INTRODUCTION

Content Based Image Retrieval (CBIR) is a prominent area in image processing due to its diverse applications in internet, multimedia, medical image archives, and crime prevention. Improved demand for image databases has increased the need to store and retrieve digital images. Extraction of visual features, viz., color, texture, and shape is an important component of CBIR [1, 20]. Out of these, Shape is one of the primary visual features in CBIR. Shape descriptors fall into two categories i.e., contour-based and region-based [3, 23]. Contour-based shape descriptors use only the boundary information by ignoring the shape interior content while region-based shape descriptors exploit interior pixels of shape [1]. Region-based shape descriptors can be applied to more general shapes. However, contour-based shape descriptors have limitations of extracting complex shapes. Hence, region based shape descriptors viz., Moment Invariants (MI) [2], Zernike Moments (ZM) [4, 5], and Legendre Moments (LM) [6] are preferred to represent the shape content of an image. An efficient shape descriptor should be affine invariant, robust, compact, easy to derive, and match [7]. The usefulness of various image moments, viz., MI, ZM, and LM as image shape features is explored for CBIR in this work.

MI is poor in the representation of image shape due to its non-orthogonality. Orthogonal moments, viz., LM and ZM can be used to represent an image with minimum amount of redundancy. Image representation using ZM has desirable properties of rotation invariance, robustness to noise, expression efficiency, effectiveness, and multilevel representation for describing various shapes of patterns [18]. Scale invariance can also be achieved with proper normalization [5]. ZM has superior performance both as region-based and shape-based descriptor but is computationally complex when compared to LM [19]. To compute the ZM of an image, the centre of the image is taken as the origin and the pixel coordinates are mapped to the range of the unit circle. Those pixels that fall outside unit circle are not used in the computation [18,19]. This is a major limitation while retrieving gray scale images because an error is introduced in discarding the pixels that fall outside the range of unit circle. This has motivated us to use Exact Legendre Moments (ELM) to represent the image effectively in this work.

Shape description or representation is an important issue both in image recognition and classification. Many techniques such as chain code, polygonal approximations, curvature, Fourier descriptors, radii method and moment descriptors have been proposed and used in various applications. Features such as moment invariants and area of region have been used in [21] but do not give perceptual shape similarity. Jain and Vailaya [22] proposed a shape representation based on the use of a histogram of edge directions. But these are not normalized to scale and also computationally expensive in similarity measures. Although Zernike moments have been viewed as the effective image descriptor, the presence of factorial calculations makes the computation of Zernike moments is complex. In [24], the authors proposed combining feature points and edges for CBIR.

Many algorithms [6,7,8] are available in the literature for the computation of LM but they are not accurate. Hosney [9] proposed a fast and accurate method for computation of Exact Legendre Moments (ELM) for gray level images. A set of 2D Legendre moments are computed exactly by using a mathematical integration of Legendre polynomials and then, a fast algorithm is applied for reduction of computation complexity. Recently, Bishnu et al. [12] proposed a method for CBIR based on SERVE (Stacked Euler Vector) i.e., Euler vector combined with Modified Moment Invariants (MMI) [15] and retrieval results are verified on COIL-20 shape database [10].

This paper is organized as follows. A review of various image moments is presented in section 2. Brief overview of the Support Vector Machine (SVM) is presented in section 3. Proposed CBIR algorithm is discussed in section 4. Experimental results are presented in section 5. Concluding remarks are given in section 6.

## 2. IMAGE MOMENTS

Image moments and their functions have been utilized as features in many image processing applications, viz., pattern recognition, image classification, target identification, and shape analysis. Moments of an image are treated as region-based shape descriptors.

### 2.1. Moment Invariants

Hu [2,17] proposed Moment Invariants (MI) for two-dimensional pattern recognition applications. Two-dimensional moments of order $(p+q)$ for digital image $f(x, y)$ is defined as follows.

$$m_{pq} = \sum_{x} \sum_{y} x^p y^q f(x, y) \qquad (1)$$

where, $p, q = 0, 1, 2 \ldots$

The summations are over the values of spatial co-ordinates $x$ and $y$ spanning the entire image. The moments in Eq. (1) are not in general invariant under translation, rotation or scale changes in the image $f(x, y)$. Translation invariance can be achieved by using central moment defined as follows.

$$\mu_{pq} = \sum \sum (x - \bar{x})^p (y - \bar{y})^q f(x, y) \qquad (2)$$

where, $\bar{x} = \dfrac{m_{10}}{m_{00}}$ and $\bar{y} = \dfrac{m_{01}}{m_{00}}$

The normalized central moment of order $(p + q)$ is defined as

$$\eta_{pq} = \frac{\mu_{pq}}{\mu_{pq}^{\gamma}} \qquad (3)$$

Where, $\gamma = \dfrac{p + q}{2} + 1$. A set of seven 2-D moment Invariants that is insensitive to translation, scale, change, mirroring, and rotation can be derived from these equations. However, the kernel is not orthogonal [2] thereby redundancy exists in image representation using these moments.

## 2.2. Zernike Moments

Zernike Moments (ZM) are orthogonal moments and can be used to represent shape content of an image with minimum amount of information redundancy [4, 5]. Orthogonal moments allow for accurate reconstruction of the image, and makes optimal utilization of shape information. Zernike Moments (ZM) are widely used in CBIR as shape descriptors [18, 19]. ZM have many desirable properties, viz., rotation invariance and robustness to noise. The complex ZM are derived by projecting the image function onto an orthogonal polynomial over the interior of a unit circle $x^2 + y^2 = 1$ as follows.

$$V_{nm}(x, y) = V_{nm}(p, \theta) = R_{nm}(\rho) \exp(jm\theta) \qquad (4)$$

$$R_{nm}(\rho) = \sum_{s=0}^{\frac{(n-|m|)}{2}} (-1)^s \frac{(n-s)!}{s! \left( \dfrac{n+|m|}{2} - s \right) \left( \dfrac{n-|m|}{2} - s \right)!} \rho^{n-2s} \qquad (5)$$

where, $n$ is non-negative integer, $m$ is an integer such that $n - |m|$ is even and $|m| < n$, $\rho = \sqrt{x^2 + y^2}$, $\theta = Tan^{-1}(\dfrac{x}{y})$. Projecting the image function onto the basis set, results Zernike moments of order $n$ with repetition $m$ given by

$$A_{nm} = \frac{n+1}{\pi} \sum_{x} \sum_{y} f(x, y) V_{nm}(\rho, \theta), \qquad (6)$$

where, $x^2 + y^2 \leq 1$.

### 2.3. Exact Legendre Moments

Legendre Moments (LM) are continuous and orthogonal moments, they can be used to represent an image with minimum amount of information redundancy. Many algorithms are developed for the computation of LM [6, 7, 8], but these methods focus mainly on 2D geometric moments. When they are applied to a digital image, a numerical approximation is necessary. Error due to approximation increases as the order of the moment increases. An accurate method for computing the Exact Legendre Moments (ELM) proposed by Hosney [9] is as follows.

Legendre moments of order $g = (p + q)$ for an image with intensity function $f(x, y)$ are defined as

$$L_{pq} = \frac{(2p+1)(2q+1)}{4} \int\limits_{-1}^{1}\int\limits_{-1}^{1} P_p(x) P_p(y) f(x, y) dx dy \tag{7}$$

where, $P_p(x)$ is the $p^{th}$ order Legendre polynomial defined as

$$P_p(X) = \sum_{k=0}^{p} a_{k,p} x^k = \frac{1}{2^p \, p!}\left(\frac{d}{dx}\right)^p \left[(x^2 - 1)\right]^p \tag{8}$$

where, $x \in [-1,1]$ and $P_p(x)$ obeys the following recursive relation

$$P_{p+1}(x) = \frac{(2p+1)}{(p+1)} x P_p(x) - \frac{p}{p+1} P_{p-1}(x) \tag{9}$$

with $P_0(x) = 1$, $P_1(x) = x$ and $P > 1$.

The set of Legendre polynomials $\{P_p(x)\}$ forms a complete orthogonal basis set on the interval [-1, 1]. A digital image of size $N \times N$ is an array of pixels. Centres of these pixels are the points $(x_i, y_j)$.

In order to improve accuracy, it is proposed to use the following approximated form [6]

$$L_{pq} = \frac{(2p+1)(2q+1)}{4} \sum_{i=1}^{N} \sum_{j=1}^{N} h_{pq}(x_i, y_j) f(x, y) \tag{10}$$

where, $x_i = -1 + \left(i - \frac{1}{2}\right)\Delta x$ and $y_j = -1 + \left(j - \frac{1}{2}\right)\Delta y$ with $i, j = 1,2,3...N$

$$h_{pq}(x_i, y_j) = \int_{x_i - \frac{\Delta x_i}{2}}^{x_i + \frac{\Delta x_i}{2}} \int_{y_j - \Delta y_j / 2}^{y_j + \Delta y_j / 2} P_p(x) P_q(y) dx dy \tag{11}$$

This double integration is required to be evaluated exactly to remove the approximation error in computation of Legendre moments. A special polynomial is given as follows.

$$\int P_p(x) dx = \frac{P_{p+1}(x) - P_{p-1}(x)}{2p+1} \tag{12}$$

where, $P \geq 1$, The set of Legendre moments can thus be computed exactly by

$$\tilde{L}_{pq} = \sum_{i=1}^{N} \sum_{j=1}^{N} I_p(x_i) I_q(y_j) f(x, y) \tag{13}$$

where,

$$I_p\left(x_i\right) = \left(\frac{(2p+1)}{(2p+2)}\right)\left[xP_p\left(x\right) - P_{p-1}(x)\right]_{U_i}^{U_{i+1}} \tag{14}$$

where,

$$I_q\left(y_j\right) = \left(\frac{(2q+1)}{(2q+2)}\right)\left[yP_q\left(y\right) - P_{p-1}(y)\right]_{V_j}^{V_{j+1}} \tag{15}$$

where,

$$U_{i+1} = x_i + \frac{\Delta x_i}{2} = -1 + i\Delta x$$

$$U_i = x_i - \frac{\Delta x_i}{2} = -1 + (i-1)\Delta x$$

similarly,

$$V_{j+1} = y_j + \frac{\Delta y_j}{2} = -1 + j\Delta y$$

$$V_j = y_j - \frac{\Delta y_j}{2} = -1 + (j-1)\Delta y$$

Equation (13) is valid only for $p \geq 1$, $q \geq 1$. Further, moment kernels can be generated using (14) and (15). Computation of ELM using (13) is time consuming.

Hence, ELM can be obtained in two steps by successive computation of 1D $q^{th}$ order moments for each row as follows. By rewriting (13) in separable form

$$\tilde{L}_{pq} = \sum_{i=1}^{N} I_p\left(x_i\right) Y_{iq} \tag{16}$$

where,

$$Y_{iq} = \sum_{i=1}^{N} I_q\left(y_j\right) f\left(x_i, y_j\right) \tag{17}$$

where, $Y_{iq}$ is the $q^{th}$ order moment of $i^{th}$ row .

Since, $I_o\left(x_i\right) = \frac{1}{N}$ . Substituting this in (16) results the following

$$\tilde{L}_{oq} = \frac{1}{N} \sum_{i=1}^{N} Y_{iq} \tag{18}$$

The number of ELM of order $g$ is given by $N_{total} = \frac{(g+1)(g+2)}{2}$ .

These ELM features are used for CBIR in this work.

## 3. SUPPORT VECTOR MACHINE

Support Vector Machines (SVMs) are supervised learning methods [14, 16] used for image classification. It views the given image database as two sets of vectors in an '$n$' dimensional space and constructs a separating hyper plane that maximizes the margin between the images relevant to query and the images not relevant to the query. SVM is a kernel method and the kernel function used in SVM is very crucial in determining the performance.

The basic principle of SVMs is a maximum margin classifier. Using the kernel methods, the data can be first implicitly mapped to a high dimensional kernel space. The maximum margin classifier is determined in the kernel space and the corresponding decision function in the original space can be non-linear [11]. The non-linear data in the feature space is classified into linear data in kernel space by the SVMs. This is illustrated in Figure 1 as follows.

The aim of SVM classification method is to find an optimal hyper plane separating relevant and irrelevant vectors by maximizing the size of the margin (between both classes).

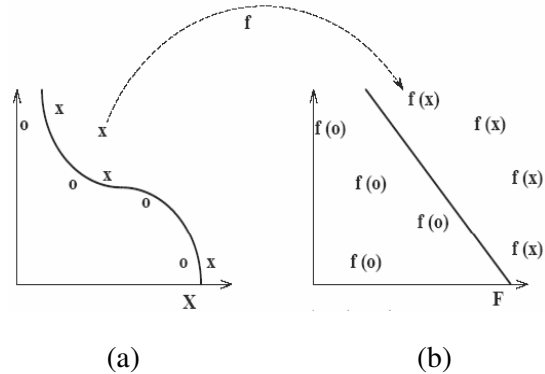

(a)                                    (b)

Figure 1. The function 'f' embeds the data in the original space

(a) kernel space (b) where the nonlinear pattern now becomes linear.

Image classification or categorization is a machine learning approach and can be treated as a step for speeding-up image retrieval in large databases and to improve retrieval accuracy. Similarly, in the absence of labelled data, unsupervised clustering is also found useful for increasing the retrieval speed as well as to improve retrieval accuracy. Image clustering inherently depends on a similarity measure, while image classification has been performed by different methods that neither require nor make use of similarity measures [13].

## 4. PROPOSED CBIR ALGORITHM

Faster and accurate CBIR algorithms are required for real time applications. This can be achieved by employing a classifier such as Support Vector Machine (SVM). SVM [11] is a supervised learning method used for image classification. It views the given image database as two sets of vectors in an '$n$' dimensional space and constructs a separating hyper plane that maximizes the margin between the images relevant to query and the images non relevant to the query. A CBIR system using ELM features and ELM features with SVM as classifier is proposed in this work.

The basic procedure involved in the proposed CBIR system is as follows.

- Computation of ELM for the given image to form the feature vector.

- Calculation of distance measure between the feature vectors of query and data base images.

- Retrieval of similar images based on minimum distance.

- Employ SVM classifier to classify the images in the database

- Increase the number of training samples to improve the classification efficiency.

The steps of proposed CBIR algorithm are as follows.

■

- Exact Legendre moments for the database images are computed by using (16)-(18) to form the feature database.

- Feature database is created by feature vector $f_{database} = (f_1, f_{2,}......f_M)$ for the image database consisting of $M$ images. Each feature vector $f_i$, for $i = 1,2....M$, is a set of ELM of order $(p + q) = g$

$$f_i = \{L_{00}L_{01}...L_{pq}\}_{database} \qquad (19)$$

- A feature vector comprising of ELM of order $(p + q) = g$ for the query image is formulated.

$$f_q = \{L_{00}L_{01}...L_{pq}\}_{query} \qquad (20)$$

- Distance measure between the feature vector $f_q$ of the query image and each feature vector of the database images $f_i$ is calculated by using Canberra distance $d_{qi}^c$.

$$d_{qi}^c = \sum_{i=1}^{M} \frac{|f_q - f_i|}{|f_q| + |f_i|} \qquad (21)$$

where, $g$ is the order of moments.

- Retrieve all the relevant images to the query image based on minimum distance $d_{qi}^c$.

- Train the SVM by selecting proper samples of the database from each class. All the classes of the image database are labelled.

- Pass the class labels with their features to the SVM classifier with the chosen kernel. The Gaussian Radial Basis Function kernel is considered as defined  in [11].

- Classify all the images from the database by considering each image in the database as the query image.

■

A query image may be any one of the database images. This query image is then processed to compute the feature vector as in equation (20). The distance $d_{qi}^c$ is computed between the query image ('$q$') and image from database ('$i$'). The distances are then sorted in increasing order and the closest sets of images are then retrieved. The top "$N$" retrieved images are used for computing the performance of the proposed algorithm. The retrieval efficiency is measured by counting the number of matches.

## 5. EXPERIMENTAL RESULTS

Retrieval performance of the proposed CBIR system is tested by conducting experiments on Corel shape database, COIL-20 [10]. It consists of 20 classes of images with each class consisting of 72 different orientations resulting in a total of 1440 images. All these gray scale images in the database are of the size 128×128. All images of all the 20 classes are used for experimentation. Experiments are conducted using MATLAB 7.2.0 with Pentium-IV, 3.00 GHz computer and osusvm toolbox.

Moments up to order 4, 5, 6, 7, 8, and 9 are considered. It results in feature vectors of dimension 9, 12, 16, 20, 25, and 30 respectively for ZM. Where as it is 15, 21, 28, 36, 45, and 55 respectively for ELM. The dimension of the feature vector is *seven* only for MI irrespective of the moment order. It is observed that the computation time is increasing with the size of the feature vector.

Comparison of the retrieval performance of the proposed CBIR system with other moment based CBIR systems is presented in Table I. Further, improved retrieval efficiency is also observed over combined SERVE and MMI [12]. Since, a classifier is employed, the average retrieval efficiency is renamed as classification efficiency. As the number of training samples increases, the classification efficiency also increases. This is presented in Table II and is illustrated in Figure 3.

TABLE I.
AVERAGE RETRIEVAL EFFICIENCY FOR VARIOUS IMAGE MOMENTS

| Method | Moment order | | | | | | |
|---|---|---|---|---|---|---|---|
| | 4 | 5 | 6 | 7 | 8 | 9 | 10 |
| Moment Invariants | 45.20 | 45.20 | 45.20 | 45.20 | 45.20 | 45.20 | 45.20 |
| Zernike Moments | 49.09 | 52.31 | 52.46 | 52.57 | 53.62 | 54.36 | 54.26 |
| Bishnu et al. [9] | 43.00 | 43.00 | 43.00 | 43.00 | 43.00 | 43.00 | 43.00 |
| Proposed (ELM) | **58.61** | **58.83** | **60.42** | **60.94** | **61.55** | **62.36** | **62.75** |

As SVM is a kernel method, the kernel function used in SVM is very crucial in determining the performance. A kernel function needs to be chosen with appropriate parameters. The kernel is tuned [14] with a pre-defined ideal kernel matrix. As a kernel method, SVMs can efficiently handle nonlinear patterns. However, the choice of kernel and tuning of appropriate parameters, adapting SVMs for specific requirements of CBIR such as learning with small sample is a challenging problem. Average retrieval times for order 9 for the CBIR systems based on MI, ZM, and ELM are 0.49, 1.25, and 0.549 seconds respectively. It is observed that MIs are faster but inefficient for CBIR. ELM based CBIR system is faster and efficient compared to other moment based CBIR systems.

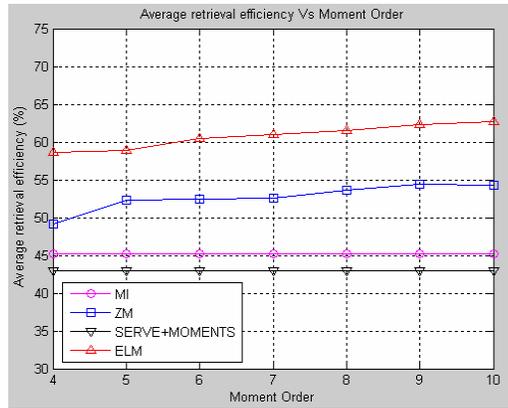

Figure 2.    Comparative Average retrieval efficiency of the proposed CBIR system for  various moments and moment orders

TABLE II.
CLASSIFICATION EFFICIENCY FOR VARIOUS IMAGE MOMENTS

| Method | Number of Training Samples | | | |
|---|---|---|---|---|
| | 4 | 5 | 6 | 7 |
| Moment Invariants | 47.71 | 48.93 | 50.64 | 51.41 |
| Zernike Moments | 82.01 | 83.26 | 84.72 | 86.74 |
| SERVE [3] (Euler+MMI) | 50.26 | 52.43 | 53.52 | 54.15 |
| Proposed | **89.51** | **91.39** | **92.57** | **94.93** |

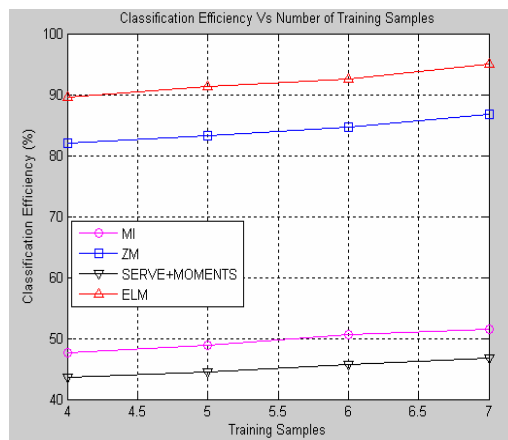

Figure 3.    Comparative Classification Efficiency of the proposed CBIR system for various moments and moment orders

## 6. CONCLUSIONS

A CBIR system using Exact Legendre Moments (ELM) is proposed in this work. Performance of the proposed CBIR system is superior compared to Bishnu et al., [12] method on COIL-20 database in terms of average retrieval efficiency and average retrieval time. It is also shown that ELMs perform better compared to other image moments, viz., MI, ZM and LM for CBIR applications.  Further, improved classification efficiency is also obtained by employing SVM classifier. It is observed that the average retrieval efficiency is increased as the moment order increases. It is also observed that the classification efficiency of the proposed CBIR system increased with the increase in the number of training samples.

**Authors**

Ch. Srinivasa rao is currently working as Professor of ECE Department, Sri Sai Aditya Institute of Science & Technology, Surampalem, AP, India. He obtained his Ph.D from University College of Engineering, JNTUK, Kakinada, A.P, India. He received M.Tech. degree from  the same institute.

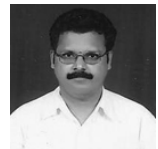

S.Srinivas kumar is currently Professor of  ECE Department, University College of Engineering and Director of Sponsored Research, JNTUK, Kakinada, AP, India. He received his M. Tech from JNTU, Hyderabad, India. He received his Ph.D. from E&EC Department, IIT, Kharagpur.

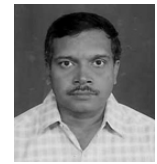

B.Chandra Mohan is currently working as Professor of ECE Department, Bapatla Engineering College, Bapatla, India. He obtained his Ph.D from JNTU College of Engineering, Kakinada, India. He obtained M.Tech from Cochin University.

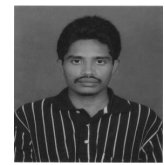